\documentclass[lettersize,journal]{IEEEtran}
\pdfoutput=1 
\usepackage{float}
 
\usepackage{cite}
\usepackage[pdftex]{graphicx}
\usepackage[section]{placeins}
\usepackage{tabularx}
\usepackage{siunitx}
\usepackage{amsmath, amssymb,mathrsfs, dsfont} 
\usepackage{amsmath,amssymb,bm,bbm}
\usepackage{graphics,graphicx}
\usepackage{multirow}
\usepackage{hyperref}
\usepackage{soul}

\usepackage{xfrac}  

\usepackage{amsmath}
\usepackage{subfiles} 
\usepackage{booktabs}

\usepackage{amsfonts}
\usepackage{booktabs}
\usepackage{siunitx}
\usepackage[dvipsnames, table]{xcolor}
\usepackage[deletedmarkup=sout,authormarkup=superscript]{changes}
\definechangesauthor[name={Mahdi}, color=NavyBlue]{MN}

\graphicspath{{\subfix{../figures/}}}

\usepackage{algpseudocode}
\usepackage{algorithm}
\usepackage{algorithmicx}
\usepackage{algcompatible}

\usepackage{subcaption}  

\usepackage{booktabs} 
\definecolor{fadedgray}{rgb}{0.8, 0.8, 0.8}  

\begin{document}
\title{Guided Multi-Fidelity Bayesian Optimization for Data-driven Controller Tuning with Digital Twins}

        
\author{Mahdi Nobar\IEEEauthorrefmark{1}, J\"urg Keller\IEEEauthorrefmark{2}, Alessandro Forino\IEEEauthorrefmark{3}, John Lygeros\IEEEauthorrefmark{4}, and Alisa Rupenyan\IEEEauthorrefmark{5}
\thanks{Maxon Motor AG and Swiss National Science Foundation supported this work through NCCR Automation, grant number 180545. (Corresponding author: Mahdi Nobar.)}
\thanks{\IEEEauthorrefmark{1} \texttt{mnobar@ethz.ch} Automatic Control Laboratory, ETH Zürich, Switzerland.}
\thanks{\IEEEauthorrefmark{2} \texttt{juerg.keller1@fhnw.ch} Automation Institute,          FHNW, Switzerland.}
\thanks{\IEEEauthorrefmark{3} \texttt{alessandro.forino@maxongroup.com} Robotic Drive Systems, Maxon Motor AG, Sachseln, Switzerland.}
\thanks{\IEEEauthorrefmark{4} \texttt{jlygeros@ethz.ch} Automatic Control Laboratory, ETH Zürich, Switzerland.}
\thanks{\IEEEauthorrefmark{5} \texttt{alisa.rupenyan@zhaw.ch} ZHAW Zurich University for Applied Sciences, ZHAW Centre for AI, Switzerland.}}



\maketitle

\begin{abstract}
We propose a \textit{guided multi-fidelity Bayesian optimization} framework for data-efficient controller tuning that integrates corrected digital twin simulations with real-world measurements.
The method targets closed-loop systems with limited-fidelity simulations or inexpensive approximations.
To address model mismatch, we build a multi-fidelity surrogate with a learned correction model that refines digital twin estimates using real data.
An adaptive cost-aware acquisition function balances expected improvement, fidelity, and sampling cost.
Our method ensures adaptability as new measurements arrive.
The digital twin accuracy is re-estimated, dynamically adapting both cross-source correlations and the acquisition function.
This ensures that accurate simulations are used more frequently, while inaccurate simulation data are appropriately downweighted.
Experiments on robotic drive hardware and supporting numerical studies demonstrate that our method enhances tuning efficiency compared to standard Bayesian optimization and multi-fidelity methods.
\end{abstract}

\IEEEpeerreviewmaketitle
\vspace{-0.5\baselineskip}
\section{Introduction}\label{sec:introduction}
In controller tuning, symbolic or physics-based simulation platforms reduce the reliance on costly real-world experiments\cite{10980269}.
Digital twins (DTs) can often capture key operational dynamics and trends, making them valuable tools for informed optimization tasks \cite{digital_twin_10947733}.
Low-cost simulations, although less accurate, offer faster and cheaper evaluation cycles compared to real experiments \cite{10499998,10874195}.
For instance, they have been successfully employed in intelligent manufacturing to accelerate optimization processes \cite{10449466, 10466425}.
DTs can enhance performance even at reduced fidelity \cite{10666830}, facilitating rapid prototyping and testing while decreasing reliance on costly physical experiments \cite{WU2026103080, 10640230}.

Multi-Fidelity Bayesian Optimization (MFBO) has been developed to leverage simulations with varying levels of fidelity and cost\cite{kennedy2000predicting, forrester2007multi}.
It incorporates multiple information sources\cite{forrester2007multi}.
MFBO balances the trade-off between high-fidelity accuracy and low-fidelity efficiency, enabling more effective optimization strategies \cite{10.1115/1.4064244}.  
For example, \cite{TAN2025112568} demonstrates MFBO in gait design for a soft quadruped, using simulation data alongside minimal physical trials to refine its controller.  
Advanced MFBO methods also account for input-dependent fidelities, where the discrepancy between low- and high-fidelity models varies with the decision variables.  
For instance, in robotics, a simulator may align well with the real system at nominal joint speeds but deviate significantly at high accelerations.
Such methods enhance data efficiency and robustness \cite{fan2024multi}, underscoring the potential of integrating additional data to improve the precision and reliability of controller tuning \cite{10836768}.

Recent works have explored online learning methods to improve control performance under changing system dynamics \cite{meng2025preserving}.
Optimization has advanced through correction models that align simulations with real-world data to create more accurate models \cite{shen2025development}.
These developments enable tuning frameworks where both the controller and a high-fidelity DT are updated during optimization, improving data efficiency and robustness to real-system changes \cite{tong2025digital, allamaa2024learning}.

We aim to improve the integration of low-fidelity, first-principles DTs with real-system measurements, increase the data efficiency of Bayesian optimization (BO) for controller tuning, and enable adaptivity to dynamic changes during optimization. 
We assume real-system data is costly, while DT simulations are cheaper but variably accurate, raising the question: \textit{How can imperfect DTs be exploited to optimize performance efficiently?}
Rather than seeking a globally accurate DT, we focus on selectively correcting it only in regions relevant to the optimization objective.
To this end, our contributions are:
\begin{itemize}
    \item A guided multi-fidelity Bayesian optimization (GMFBO) framework that integrates real and simulated data for data-efficient controller tuning.
    \item A correction mechanism that refines DT predictions where it matters for optimization and quantifies DT accuracy.
    \item An adaptive surrogate model that adjusts cross-source correlations based on DT accuracy.
    \item A cost-aware acquisition function that balances fidelity, expected improvement, and sampling cost, mitigating overreliance on inaccurate simulations.
\end{itemize}
We validate GMFBO on a real robotic system, including scenarios with changing friction, demonstrating substantial gains in tuning efficiency over baseline methods.

\vspace{-0.5\baselineskip}
\section{Controller Tuning Problem}\label{sec:Controller_Tuning_Problem}
\begin{figure}[!t] 
\centering
\vspace{0.3\baselineskip}
\includegraphics[width=.99\columnwidth, trim=3.7cm 0.1cm 3.7cm 0cm, clip]{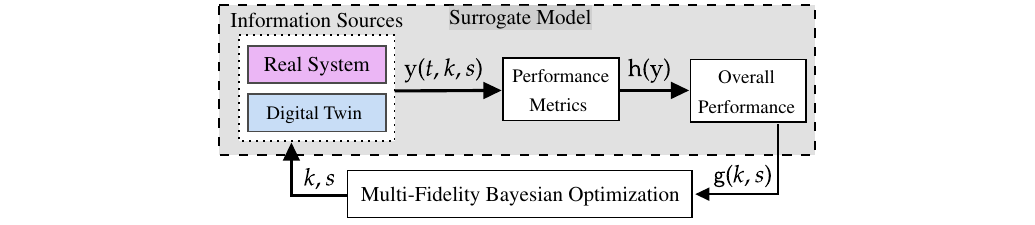}
\caption{Efficient MFBO controller tuning using multiple information sources}
\vspace{-1.5\baselineskip}
\label{fig:problem_schematic}
\end{figure}
We consider two information sources that provide data for controller tuning.
The \textit{target information source} (IS1) is a closed-loop system comprising the plant, controller, and associated components.
We make no assumptions about the parametric controller and plant structures.
Measurements from IS1 are accurate up to measurement noise.
An additional information source (IS2) serves as the system’s \textit{digital twin}, which is less accurate than IS1.
By accuracy (or fidelity), we mean the agreement between IS2’s predicted objective and IS1’s measured objective under identical controller parameters.
In many practical scenarios, the digital twin—constructed using physics-based modeling, empirical data, or a combination thereof—cannot be perfectly identified due to limited access to internal states or safety constraints.

Let $k \in \mathcal{K}$ be the controller parameter vector within a bounded box $\mathcal{K} \subseteq \mathbb{R}^{n_k}$. 
The set $\mathcal{K}$ is restricted to gains that ensure a stable, non-divergent unit step response on both the real system and the digital twin over the horizon $T$, thereby excluding unsafe controllers.
We define $s \in \mathcal{S} = \{0,1\}$ as a binary fidelity variable correlated with source accuracy, where $s = 1$ denotes the high-fidelity IS1 and $s = 0$ the low-fidelity IS2.
The augmented input random variable is $z:=[k,s]$ where $z \in \mathcal{K} \times \mathcal{S}$, with dimension $n_{k}+1$.
Let $\hat{\mathsf{g}}: \mathcal{K} \times \mathcal{S} \to \mathbb{R}$ denote the information source–dependent objective function.
There is no assumption on the convexity of the objective function.

We evaluate an information source $s$ at a given controller parameter $k$, measuring the feedback system \textit{output} $\mathrm{y}(t,k,s) \in \mathbb{R}$, given the reference input signal $\mathrm{y}^{*}(t) \in \mathbb{R}$ at time $t \in [0,T]$.
Performance metrics are $h_{i}(\mathrm{y}(t,k,s)) \in \mathbb{R}$ for $i=1,\ldots,n_{\mathrm{h}}$.
These metrics quantify various aspects of system behavior.
We adopt a weighted sum approach to convert the multi-objective problem into a single-objective formulation \cite{pereira2022review}, assigning relative importance to each performance metric \cite{5429562}.
So the overall performance cost function $\hat{f}(k)$ is \vspace{-0.2\baselineskip}
\begin{equation}\label{eq:f_hat}
\hat{\mathsf{g}}(k, s) := \mathrm{w}^\mathsf{T} \mathrm{h}(\mathrm{y}(t,k,s))=\sum_{i=1}^{n_{\mathrm{h}}}w_{i}h_{i}(\mathrm{y}(t,k,s)),
\vspace{-0.2\baselineskip}
\end{equation}
where $\mathrm{h}(\mathrm{y}):=[h_{1}(\mathrm{y}),...,h_{n_{\mathrm{h}}}(\mathrm{y})]$ and $\mathrm{w}:=[w_{1},...,w_{n_{\mathrm{h}}}]$.
The weights vector $\mathrm{w}$ reflects the priority and scale of our metrics.
Figure \ref{fig:problem_schematic} visualizes the problem setup.

Due to measurement noise, an observation $\mathsf{g}$ at input $z$ is modeled as
\vspace{-0.2\baselineskip}
\begin{equation}\label{eq:measurement_noise}
    \mathsf{g}(z) = \hat{\mathsf{g}}(z) + \eta,
\vspace{-0.2\baselineskip}
\end{equation}
where $\eta \sim \mathcal{N}(0, \sigma_{\eta}^2)$ denotes zero-mean Gaussian noise with variance $\sigma_{\eta}^2$.
This assumption is widely used because it effectively captures a broad range of independent random error sources, as per the central limit theorem \cite{thrun_10.5555/1121596}.
Our \textit{data-driven controller tuning problem} is to obtain the optimum controller parameter vector $k^{*}$ for IS1 by solving
\vspace{-0.3\baselineskip}
\begin{equation}\label{eq:k_star}
\begin{split}
k^{*}:=\arg\min_{k \in \mathcal{K}}\ \hat{\mathsf{g}}(k,1).
\end{split}
\vspace{-0.2\baselineskip}
\end{equation}
Evaluating the objective function $\mathsf{g}(k,1)$ is expensive, since it involves operating the real setup.

\vspace{-0.5\baselineskip}
\section{Guided Multi-fidelity Bayesian Optimization}\label{sec:Guided_BO}

\begin{figure}[!h] 
\centering
\vspace{-0.2\baselineskip}
\includegraphics[width=.999\columnwidth, trim=2.85cm 0.1cm 2.85cm 0.05cm, clip]{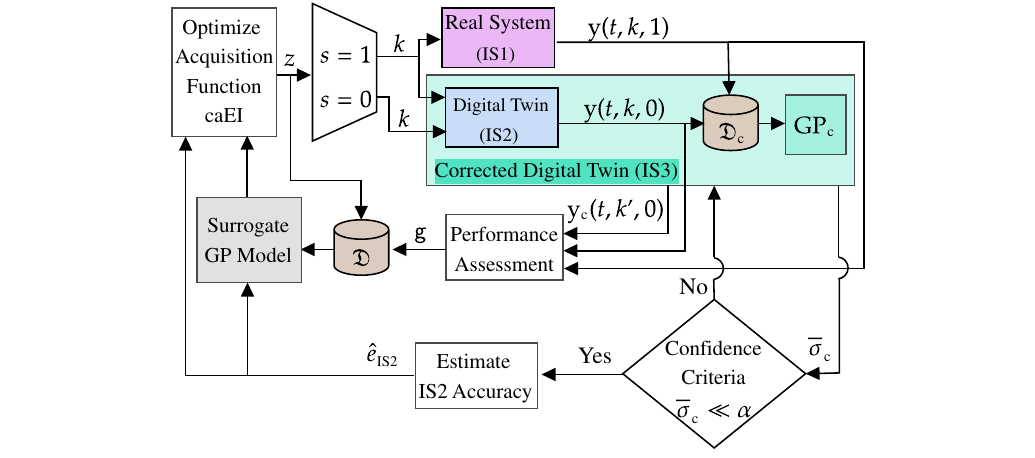}
\caption{Guided multi-fidelity Bayesian optimization for tuning controller parameters}\vspace{-1\baselineskip}
\label{fig:GMFBO_schematic}
\end{figure}

Figure~\ref{fig:GMFBO_schematic} illustrates the information flow and interaction between the different sources in GMFBO. 
Data from IS1 is used both to correct the DT as an additional information source (IS3) and to provide high-fidelity observations for BO.
The acquisition function selects the information source by incorporating the estimated DT mismatch (via the error metric $\hat{e}_{\text{IS2}}$) at each candidate.
Most performance evaluations are delegated to the corrected DT in IS3, reducing the need for expensive IS1 queries.
We next describe the surrogate objective model, the corrected source IS3 and its GP-based DT correction, the estimation of DT accuracy, the adaptive kernel and acquisition function, and finally the GMFBO algorithm.

\vspace{-0.5\baselineskip}
\subsection{Multi-fidelity Surrogate Model}\label{subsec:Improved_Multi-fidelity_Bayesian_Optimization}
We define a \textit{surrogate} Gaussian process to model the multi-fidelity objective $\hat{\mathsf{g}}(z)$.
A key contribution of our approach is to treat the fidelity variable $s$ as an explicit input to the GP, enabling a unified surrogate that seamlessly integrates data from all sources and implicitly weighs them according to their fidelity.
We assume a GP prior over the function values $\mathbf{g}(\mathbf{z}) \sim \mathrm{GP}(\mathbf{m}(\mathbf{z}), \mathbf{c}(z))$, where $\mathbf{g}(\mathbf{z}) := [\mathsf{g}(z_{1}), \dots, \mathsf{g}(z_{n})]^\top$ represents the vector of function values at specific training points $\mathbf{z} := [z_{1}, \dots, z_{n}]$, prior means $\mathbf{m}(\mathbf{z}) := [m(z_{1}), \dots, m(z_{n})]^\mathsf{T}$, prior covariances $ \mathbf{c}(z) := [c(z,z_{1}), \dots, c(z, z_{n})]^\mathsf{T}$, $n$ is the number of observed points, $\mathbf{m}$ and $\mathbf{c}$ are the prior mean and covariance functions, respectively.
Define initial data set 
\vspace{-0.2\baselineskip}
\begin{equation}\label{eq:D_init}
    \mathfrak{D}_{\text{init}}:=\bigl\{ \bigl(z_i, \mathsf{g}(z_i)\bigl)\: |\: i\in\{1,...,N_{0_{s=1}}+N_{0_{s=0}}\} \bigl\},
\vspace{-0.2\baselineskip}
\end{equation}
where $N_{0_{s=1}}$ is the initial IS1 data size and $N_{0_{s=0}}$ the initial IS2 data size.
The likelihood function for $n$ observations is then $p(\mathbf{g} |  \mathbf{z}, \hat{\mathbf{g}}) = \mathcal{N}(\hat{\mathbf{g}}, \mathbf{\Sigma})$, where $\mathbf{\Sigma} = \text{diag}(\sigma_{\eta_{1}}^2, \dots, \sigma_{\eta_{n}}^2)$ is a diagonal matrix of observational noise variances.
The hyperparameters of the covariance kernel and observation noise are learned by maximizing the log marginal likelihood \cite{williams2006gaussian}.
We define \textit{surrogate training dataset} $\mathfrak{D} \in \mathbb{R}^{n \times (n_{k}+2)}$ with $n$ data points 
\begin{equation}\label{eq:D}
\mathfrak{D}:=\bigl\{ \bigl(z_i, \mathsf{g}(z_i)\bigl)\: |\:  i \in\{1,...,n\} \bigl\}.
\vspace{-0.2\baselineskip}
\end{equation}
The predictive distribution of $\hat{\mathsf{g}}(z_*)$ at a test point $z_*$ is $\hat{\mathsf{g}}(z_*) | \mathfrak{D} \sim \mathcal{N}(\mu_*, \sigma_*^2)$, where the predictive mean and variance are respectively $\mu_* := m(z_*)+\mathbf{c}(z_*)^T (\mathbf{C} + \mathbf{\Sigma})^{-1} (\mathbf{g}(\mathbf{z})-\mathbf{m}(\mathbf{z})),$ and $\sigma_*^2 := c(z_*, z_*) - \mathbf{c}(z_*)^T (\mathbf{C} + \mathbf{\Sigma})^{-1} \mathbf{c}(z_*),$
with  $\mathbf{c}(z_*)$ being the covariance vector between $z_*$ and the training data, and $\mathbf{C}:=[c({z}_{\xi_{1}},{z}_{\xi_{2}})]\bigr\rvert_{\xi_{1},\xi_{2}=1}^{n}\in \mathbb{R}^{n \times n}$ \cite{williams2006gaussian}.

\vspace{-0.5\baselineskip}
\subsection{Digital Twin Correction}
Our proposed approach focuses on improving the digital twin's estimation of the performance cost function via a learned correction. The corrected digital twin $\mathrm{GP}_{\text{c}}$ is a Gaussian process regression model that predicts the real system output $\hat{y}(t, k, s=1)$ at each time step $t$ using the simulated output $y(t, k, s=0)$ from IS2 at that time step. $\mathrm{GP}_{\text{c}}$ is trained on a dataset of paired observations: $ \mathfrak{D}_{\text{c}} := \{ ([t, \mathrm{y}(t,k,s=0)], \mathrm{y}(t,k,s=1)) \mid t \in [0,T] \}$, with Gaussian observation noise,
        $\mathrm{y}(t,k,s=1) = \hat{\mathrm{y}}(k,t,s=1) + \epsilon$,
where $\epsilon \sim \mathcal{N}(0,\sigma_{\epsilon}^2)$ is the output measurement noise.
Before training, inputs and targets are normalized to zero mean and unit variance to support stable hyperparameter learning \cite{matlab_fitrgp}.

The correction model enables efficient data augmentation: each real evaluation at controller gains $k$ yields $n_c$ locally corrected samples around $k$.
We set the total number of BO iterations on IS1 and IS2 as $N$.
Each time the optimizer samples IS1 at controller gains $k$, we draw $k' \sim \mathcal{N}(k,\,0.05\,\mathbb{K}) \in \mathcal{K}$, where $\mathbb{K}$ is diagonal with entries equal to the feasible parameter ranges in $\mathcal{K}$.
The factor $0.05$ sets the per-parameter standard deviation to $5\%$ of its range (elementwise, since $\mathbb{K}$ is diagonal), localizing IS3 evaluations near the latest IS1 point.
We found that using a factor between $0.02-0.1$ yields similar behavior.
We draw $n_{\text c}$ corrected IS3 samples to the surrogate dataset $\mathfrak{D}$, per one real-system evaluation.
If the draw $k'$ falls outside $\mathcal{K}$, we project it back onto $\mathcal{K}$.

Then, IS2 generates a sequence of estimated outputs $\mathrm{y}(t,k',0)$ for $t \in [0,\ldots,T]$. The sequence is processed point-by-point through $\mathrm{GP}_{\text{c}}$, and provides the corrected IS1 output at each time step, denoted by
\vspace{-0.2\baselineskip}
\begin{align}
    \mathrm{y}_{\text{c}}(t,k',0) := \mu_{\text{c}}([t, \mathrm{y}(t,k',0)]).
\vspace{-0.2\baselineskip}
\end{align}
The standard deviation of this prediction is then denoted by $ \sigma_{\text{c}}([t, \mathrm{y}(t,k',0)])$.
We define the prediction uncertainty as
\vspace{-0.2\baselineskip}
\begin{equation}\label{eq:bar_sigma_c}
    \bar{\sigma}_{\text{c}} := \sqrt{\frac{1}{T} \sum_{t=0}^{T} \sigma^2_{\text{c}}([t, \mathrm{y}(t,k',0)])},
\vspace{-0.2\baselineskip}
\end{equation}
which measures the average predictive standard deviation of $\mathrm{GP}_{\text{c}}$ across the trajectory and quantifies the confidence of the correction model in its predictions.  
We impose the condition
\vspace{-0.2\baselineskip}
\begin{equation}\label{eq:IS3_ki_acceptance}
\bar{\sigma}_{\text{c}} \ll \alpha:= \sqrt{\frac{1}{T} \sum_{t=0}^{T} (\mathrm{y}^{*}(t))^2} \in \mathbb{R},
\vspace{-0.2\baselineskip}
\end{equation}
where $\alpha$ is the root-mean-square (RMS) of the reference signal, providing a characteristic scale for evaluating prediction uncertainty.  
If \eqref{eq:IS3_ki_acceptance} holds, the corrected prediction is deemed sufficiently confident, the data point is accepted and we update $\mathfrak{D}$: $
\mathfrak{D} \gets \mathfrak{D} \cup \{[k', 0], \mathsf{g}_{\text{c}}(k', 0)\}$, 
where $\mathsf{g}_{\text{c}}(k', 0)$ is the performance estimated using IS3.
For each IS1 query, this procedure is repeated to generate $n_{\text c}$ accepted IS3 samples that are added to $\mathfrak{D}$. 
 For a unit step reference $\mathrm{y}^{*}(t)=1$ for $t\geq 0$, we obtain $\alpha=1$. Here, $\alpha$ equals $100\%$ of the normalized reference output, so $\bar{\sigma}_{\text{c}} \ll 1$ ensures the model’s uncertainty remains small relative to maximum expected magnitude. 

Adding several corrected points around each real measurement increases the local accuracy of the surrogate model, and reveals which directions in parameter space lead to improved performance.
By providing this information, IS3  guides the optimizer towards promising regions of the objective landscape.

Each time $\mathsf{g}_{\text{c}}(k',0)$ is estimated by IS3, the corresponding $\mathrm{y}(t,k',0)$ samples from IS2 are already available, so computing $\mathsf{g}_{\text{c}}(k',0)$ incurs no additional cost.
We quantify the mismatch of IS2 relative to IS3 by
\vspace{-0.2\baselineskip}
\begin{equation}\label{eq:e_IS2}
    \hat{e}_{\text{IS2}}
    := \frac{1}{N_{\text{c}}}\sum_{i=1}^{N_{\text{c}}}
       \frac{\lvert \mathsf{g}_{\text{c}}(k'_{i},0) - \mathsf{g}(k'_{i},0) \rvert}
            {\lvert \mathsf{g}_{\text{c}}(k'_{i},0) \rvert},
\vspace{-0.2\baselineskip}
\end{equation}
where $N_{\text{c}}$ is the total number of IS3 samples in $\mathfrak{D}$.
Large $\hat{e}_{\text{IS2}}$ indicates that IS2 predictions deviate from the corrected estimates, suggesting low simulation fidelity.

\vspace{-0.7\baselineskip}
\subsection{Adaptive Multi-Source Kernel Function}
We describe how different data sources are coupled within the multi-fidelity GP surrogate model for the BO objective.
Building on the kernel formulation proposed for multi-fidelity GP models in \cite{poloczek2017multi, kernel_10.5555/3692070.3693777}, we define the composite surrogate kernel for the BO optimization objective model as
\vspace{-0.1\baselineskip}
\begin{equation}\label{eq:kernel}
    c(z_1, z_2) := \gamma_{0}(s_1, s_2,\hat{e}_{\text{IS2}})\, c_0(k_1, k_2) + \gamma_{1}(s_1, s_2)\, c_1(k_1, k_2),
\vspace{-0.4\baselineskip}
\end{equation}
with components
\vspace{-0.2\baselineskip}
\begin{align}
    c_i(k_1, k_2) &:= \sigma_i^2 \, \mathcal{M}\!\left(\|k_1 - k_2\|, l_i\right), \quad i \in \{0,1\}, \label{eq:c_i} \\
    \gamma_{0}(s_1, s_2,\hat{e}_{\text{IS2}}) &:= \mathcal{M}\!\left(\|s_1 - s_2\|, l_{\gamma_0}(\hat{e}_{\text{IS2}})\right), \label{eq:gamma_0} \\
    \gamma_{1}(s_1, s_2) &:= (1 - s_{1})(1 - s_{2})(1 + s_{1} s_{2})^{p}, \nonumber
\vspace{-0.2\baselineskip}
\end{align}
where $\mathcal{M}(r,l) := \left(1 + \tfrac{\sqrt{5}r}{l} + \tfrac{5r^2}{3l^2}\right)\exp\!\left(-\tfrac{\sqrt{5}r}{l}\right)$ is the Matérn-5/2 kernel, $\sigma_i^2$ are output variances, and $p$ modulates correlation among sources with $s<1$.
The hyperparameters $l_i$, $\sigma_i^2$, and $p$ are learned via MAP estimation by maximizing the marginal log-likelihood with log-priors on $l_i$ \cite{williams2006gaussian}.  
The mismatch metric $\hat{e}_{\text{IS2}}$ from \eqref{eq:e_IS2} adjusts $\gamma_{0}$ to reduce cross-fidelity correlation when IS2 accuracy degrades.
Here, $c_0(k_1,k_2)$ captures correlation between IS1 (high fidelity) and IS2 or IS3 (low fidelity) as a function of controller-parameter distance, while $\gamma_{1}$ captures correlations between IS2 and IS3 data (both low fidelity sources), and multiplication by $c_1(k_1,k_2)$ further incorporates controller-parameter similarity.  
Overall, the kernel jointly accounts for distances in controller space and fidelity-dependent correlations across all information sources.

To \textit{adapt} the IS2–IS1 correlation to the estimated IS2 accuracy, we use the clamping operator
$\mathcal{P}_{[a,b]}(x) := \min\{b, \max\{a, x\}\}$, which projects $x \in \mathbb{R}$ into $[a,b]$.  
The cross–source lengthscale is defined as
\vspace{-0.2\baselineskip}
\begin{equation}\label{eq:l_gamma_0_clamp}
    l_{\gamma_{0}}(\hat{e}_{\text{IS2}}) := \mathcal{P}_{[l_{\gamma_{0_{\text{min}}}},\,l_{\gamma_{0_{\text{max}}}}]}\!\left(\frac{1}{\hat{e}_{\text{IS2}}}\right),
\vspace{-0.2\baselineskip}
\end{equation}
where $\hat{e}_{\text{IS2}}$ is the mismatch in \eqref{eq:e_IS2}.
The clamp bounds $[l_{\gamma_{0_{\text{min}}}},\,l_{\gamma_{0_{\text{max}}}}]:=[0.1,\,1]$ prevent under- or overfitting while keeping cross-source correlation responsive to $\hat{e}_{\text{IS2}}$.
Large $\hat{e}_{\text{IS2}}$ (poor DT accuracy) pushes $l_{\gamma_{0}}$ toward $0.1$ (near-zero correlation), whereas small $\hat{e}_{\text{IS2}}$ (high DT accuracy) drives it toward $1$, enabling moderate correlation while preventing overreliance on simulation data.

\vspace{-0.5\baselineskip}
\subsection{Adaptive Cost-aware Acquisition Function} 

The information source for the next measurement is selected by extending the expected improvement (EI) acquisition to incorporate source-dependent sampling cost.
We define the \textit{fidelity-aware sampling cost} as
\vspace{-0.2\baselineskip}
\begin{equation}\label{eq:H_samplng_cost}
\mathcal{H}(s|\hat{e}_{\text{IS2}}) :=
\begin{cases}
1, & s = 1, \\[4pt]
\mathcal{P}_{[\mathcal{H}_{\text{min}},\,\mathcal{H}_{\text{max}}]}\bigl(\hat{e}_{\text{IS2}}\bigr), & s = 0.
\end{cases}
\vspace{-0.3\baselineskip}
\end{equation}
 IS1 evaluations always have unit cost, while the cost for IS2 evaluations increases with the mismatch $\hat{e}_{\text{IS2}}$, with clamping to $[\mathcal{H}_{\text{min}},\,\mathcal{H}_{\text{max}}]:=[0.1,1]$ ensuring numerical stability.
We then define a \textit{cost-aware Expected Improvement acquisition function} (caEI) as 
\vspace{-0.2\baselineskip}
\begin{equation}\label{eq:caEI}
\text{caEI}(z|\hat{e}_{\text{IS2}}):=\frac{a_{\text{EI}}(z)}{\mathcal{H}(s|\hat{e}_{\text{IS2}})},
\vspace{-0.2\baselineskip}
\end{equation}
where $a_{\text{EI}}(z)$ is the EI acquisition function with a closed-form formulation \cite{pmlr-v222-zhou24a}. It is computed independently for each information source such that
\vspace{-0.2\baselineskip}
\begin{equation}\label{eq:EI}
a_{\text{EI}}(z):=\sigma(z)\bigl(\upsilon(z)\Phi(\upsilon(z))+\varphi(\upsilon(z))\bigr),
\vspace{-0.2\baselineskip}
\end{equation}
with $\upsilon(z):=(\mu(z)-\mathsf{g}(s)^{+})\sigma(z)^{-1}$, 
where $\mu(z)$ and $\sigma(z)$ are the surrogate GP posterior mean and standard deviation, $\mathsf{g}(s)^{+}$ is the best observed objective in fidelity $s$, $\Phi(\cdot)$ is the standard normal cdf, and $\varphi(\cdot)$ is the pdf.  
When DT fidelity is high (small mismatch $\hat{e}_{\text{IS2}}$), the denominator in \eqref{eq:caEI} decreases, encouraging DT sampling due to increasing $\text{caEI}$. When  $\hat{e}_{\text{IS2}}$ is high, the cost approaches 1, discouraging reliance on inaccurate simulations. IS3 points are added after IS1 is selected, not through the acquisition function.
Clamping avoids degenerate cases: the lower bound prevents division by a minimal denominator, and the upper bound ensures DT queries are never more expensive than real-system ones.  
This formulation enables the optimizer to adaptively switch between IS1 and IS2 based on the current accuracy of the DT.

\vspace{-0.5\baselineskip}
\subsection{Guided Multi-Fidelity BO Algorithm}
Our GMFBO framework leverages real data to refine the DT and provide more accurate performance estimates to \textit{guide} the optimizer. The DT is selectively corrected in regions of the parameter space that are relevant for reaching the optimization objective.  
This distinguishes our approach from classical active learning, which focuses on improving model accuracy, aiming to enhance predictive fidelity everywhere.  

Algorithm~\ref{alg:GMFBO} summarizes the procedure.  
Real-system measurements are used to estimate the mismatch \eqref{eq:e_IS2}, which in turn (i) adapts $l_{\gamma_0}$ in \eqref{eq:l_gamma_0_clamp} and $\mathcal{H}$ in \eqref{eq:H_samplng_cost} to adjust the influence of simulated data, and (ii) updates the kernel in \eqref{eq:kernel} to capture evolving fidelity-dependent correlations between real and simulated observations.  
At the acquisition level, the correction model modifies the sampling cost $\mathcal{H}$ in \eqref{eq:H_samplng_cost}, enabling the optimizer to balance exploration of the DT with the cost and reliability of real experiments.  

In contrast to standard MFBO, which passively exploits correlations across fixed fidelities, GMFBO actively reshapes these correlations based on online mismatch estimates and focuses DT refinement only where it matters for optimization.  
This mechanism provides a direct pathway by which the DT and correction model \textit{guide} the optimizer, as demonstrated in the following sections.

\vspace{-0.2\baselineskip}
\begin{algorithm}[!h]
\caption{Guided Multi-Fidelity Bayesian Optimization}\label{alg:GMFBO}
\begin{algorithmic}[1]
\STATE \textbf{Initialize} $\mathcal{K}$, $N$, $n_{\text{c}}$, $\mathfrak{D}_{\text{c}}$, $\mathrm{GP}_{\text{c}}$, $\mathfrak{D} \gets \mathfrak{D}_{\text{init}}$, $\hat{e}_{\text{IS2}}$, $\alpha$
\FOR{BO iteration $n = 1, \dots, N$}
    \STATE Train surrogate GP model given $\mathfrak{D}$ 
    \STATE Select candidate $(k, s) \gets \arg\underset{z{=}[k,s]}{\max}\,\text{caEI}_n(z|\hat{e}_{\text{IS2}})$ 
    \vspace{-0.2\baselineskip}
    \IF{ $s=1$ }
        \STATE Evaluate real system: $(t, \mathrm{y}(t,k,1), \mathsf{g}(k, 1)) \leftarrow \text{IS1}(k)$
        \STATE Update $\mathfrak{D} \gets \mathfrak{D} \cup \{[k, 1], \mathsf{g}(k,1)\}$ 
        \STATE Evaluate DT: $\mathrm{y}(t,k,0) \leftarrow \text{IS2}(k)$ 
        \STATE Update $\mathfrak{D}_{\text{c}} \gets \{([t, \mathrm{y}(t,k,0)], \mathrm{y}(t,k,1))\,|\, t \in [0,T]\}$ 
        \STATE Update correction model $\mathrm{GP}_{\text{c}}\;,\;i\leftarrow 0$
        \WHILE{$i<n_{\text{c}}$}
            \STATE Sample $k'\sim \mathcal{N}(k,\,0.05\,\mathbb{K}) \in \mathcal{K}$
            \STATE Evaluate corrected DT: $\mathsf{g}_{\text{c}}(k',0)  \leftarrow \text{IS3}(k')$ 
            \STATE Compute $\bar{\sigma}_c$ from \eqref{eq:bar_sigma_c}
            \IF{$\bar{\sigma}_{\text{c}} \ll \alpha$} 
            \STATE Update $\mathfrak{D} \gets \mathfrak{D} \cup \{[k', 0], \mathsf{g}_{\text{c}}(k',0)\}\;,\;i\leftarrow i+1$
            \ENDIF
        \ENDWHILE
        \State Recompute the mismatch $\hat{e}_{\text{IS2}}$ from \eqref{eq:e_IS2}
        \State Update $l_{\gamma_{0}}$ \eqref{eq:gamma_0} and $\mathcal{H}$ \eqref{eq:H_samplng_cost} based on new $\hat{e}_{\text{IS2}}$
    \ELSIF{$s=0$}
        \STATE Compute DT performance: $\mathsf{g}(k,0) \gets \text{IS2}(k)$ 
        \STATE Update $\mathfrak{D} \gets \mathfrak{D} \cup \{[k,0], \mathsf{g}(k,0)\} $ 
    \ENDIF
\ENDFOR
\STATE \textbf{return} $k^* \gets \arg\min\limits_{k} \mathsf{g}(k,s=1)\,,\: \forall (k,1)\in \mathfrak{D}$ 
\end{algorithmic}
\end{algorithm}
\vspace{-0.5\baselineskip}
\vspace{-0.5\baselineskip}
\section{System Modelling}\label{sec:System_Modelling}
We use the high-efficiency joint (HEJ) 90 drive system from maxon, and iteratively tune the parameters of its joint position-velocity-torque (JPVT) controller.
The system comprises a high-torque permanent magnet synchronous motor, a planetary gearbox, dual encoders, and power electronics, with an EPOS4 controller that features a hierarchical structure consisting of a low-level current loop and higher-level position, velocity, and torque loops.  
The JPVT controller uses a proportional–derivative position strategy with tunable parameters $k=(K_p,K_d)$, constrained by
\vspace{-0.1\baselineskip}
\begin{equation}
    \mathcal{K} = \{ (K_p, K_d) \mid K_p \in [K_{p_\text{min}}, K_{p_\text{max}}],\; K_d \in [K_{d_\text{min}}, K_{d_\text{max}}] \},
\vspace{-0.1\baselineskip}
\end{equation}
with $(K_{p_\text{min}}, K_{p_\text{max}}, K_{d_\text{min}}, K_{d_\text{max}})=(30,200,2,10)$ for numerical analysis and $(70,120,2,5)$ for real experiments.  
Safety within $\mathcal{K}$ can be enforced during optimization by incorporating barrier functions into $\mathsf{g}$ \cite{AR_MK_saftyaware} or adopting safe exploration strategies \cite{10288044}.

\begin{figure}[!h]
\centering
\vspace{-0.5\baselineskip}
\begin{subfigure}[b]{0.85\columnwidth}
    \centering
    \includegraphics[width=\linewidth, trim=3.2cm 0.1cm 3.2cm 0.1cm, clip]{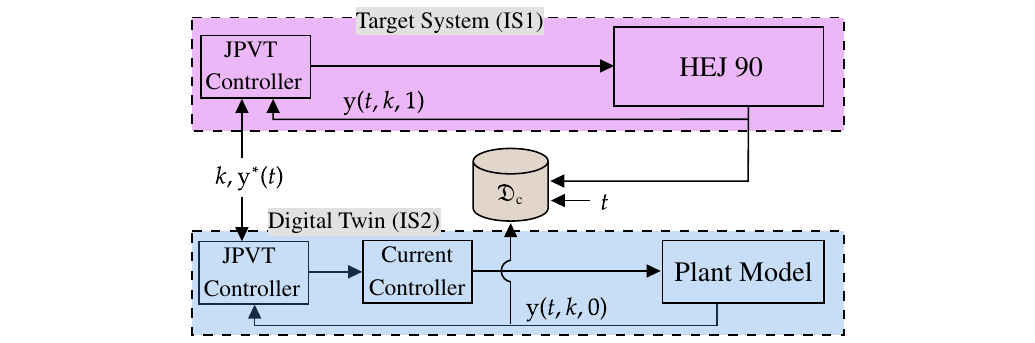}
    \caption{Block diagram of IS1 and IS2 with correction data set}
    \label{fig:Dc_dataset}
\end{subfigure}
\vspace{1.5em} 
\begin{subfigure}[b]{0.85\columnwidth}
    \centering
    \includegraphics[width=\linewidth, trim=2.75cm 0.1cm 2.5cm 0cm, clip]{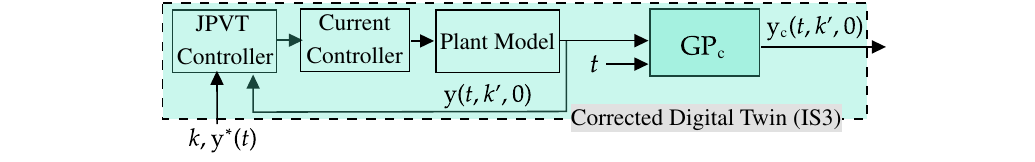}
    \vspace{-1.5\baselineskip}
    \caption{Block diagram of IS3 \label{fig:IS3}}
\end{subfigure}
\vspace{-1.5\baselineskip}
\caption{Block diagrams of information sources}
\label{fig:block_diagrams_mismtach_all}
\vspace{-0.5\baselineskip}
\end{figure}

Figure \ref{fig:Dc_dataset} shows the information-source block diagram.  
Our objective is a weighted sum of four performance metrics from the position step response.  
A \textit{unit} step reference $\mathrm{y}^{*}(t)=0$ for $t<t_{0}=0.1$ [s] and $\mathrm{y}^{*}(t)=1$ [rad] ($652$ encoder counts) for $0 \leq t-t_{0} \leq T=1$ [s] is applied to the motor output, yielding $\alpha=1$ from \eqref{eq:IS3_ki_acceptance}.
We define $\mathrm{h}(k)=[O_{\text{s}},T_{\text{tr}},T_{\text{r}},T_{\text{s}}]$, where $O_{\text{s}}$, $T_{\text{tr}}$, $T_{\text{r}}$, and $T_{\text{s}}$ denote overshoot, transient, rise, and settling time, respectively.
The objective function $\mathsf{g}$ for augmented input $z$ is
\vspace{-0.2\baselineskip}
\begin{equation}\label{eq:g}
    \mathsf{g}(z) = w_1 \cdot O_{\text{s}} + w_2 \cdot T_{\text{tr}} + w_3 \cdot T_{\text{r}} + w_4 \cdot T_{\text{s}},
\vspace{-0.2\baselineskip}
\end{equation}
with weights $w_{i},\, i=1,\ldots,4$.
Since $\hat{e}_{\text{IS2}}$ in \eqref{eq:e_IS2} is based on these metrics, the mismatch mainly reflects transient step-response errors, while effects outside this window or uncaptured by the metrics (e.g., backlash) may remain unobserved.
To construct $\mathfrak{D}_{\text{init}}$, we randomly sample $(K_p,K_d)$ pairs from $\mathcal{K}$ and record the resulting $\mathrm{h}(k)$.
The $\mathrm{GP}_{\text{c}}$ model is retrained whenever $\mathfrak{D}_{\text{c}}$ is updated.  
Figure \ref{fig:IS3} illustrates IS3, where the learned model corrects DT outputs.

We use a digital twin of the drive system (IS2), based on a nonlinear inverse flux linkage motor model \cite{en12050783} refined to account for friction, ripple harmonics, and a simple gearbox backlash model \cite{matlabGearBacklash}.  
Table \ref{tab:GMFBO_params} summarizes the chosen values for the design parameters of our method.

\begin{table}[!h]
\centering
\vspace{-0.4\baselineskip}
\caption{Design Parameters of the GMFBO Method}
\label{tab:GMFBO_params}
\vspace{-0.3\baselineskip}
\renewcommand{\arraystretch}{1.2}
\begin{tabular}{l l l l}
\hline
\textbf{Parameter} & \textbf{Expected Range} & \textbf{Chosen Value} & \textbf{Reference} \\
\hline
$N_{0_{s=1}}$    & $\geq1$     & $2$    & Subsection \ref{subsec:baseline_comparison} \\
$N_{0_{s=0}}$     & $\geq1$     & $10$    & Subsection \ref{subsec:baseline_comparison}  \\
$N$              & $\geq1$     & $20$   & Algorithm $\ref{alg:GMFBO}$ \\
$n_{\text{c}}$   & $\geq1$     & $4$    & Algorithm $\ref{alg:GMFBO}$ \\
$\alpha$         & $\in [0,1]$        & $1$           & \eqref{eq:IS3_ki_acceptance} \\
\hline
\end{tabular}
\vspace{-0.5\baselineskip}
\end{table}

\vspace{-0.5\baselineskip}
\section{Numerical Analysis}\label{sec:Numerical_Analysis}
For the numerical analysis, IS1 is the original high-fidelity DT model from maxon with a lookup table mapping voltage to current in the non-linear motor model.  
IS2 is identical, except that Gaussian noise with zero mean and standard deviation equal to $50\%$ of the respective data range is added to the lookup table, reducing its accuracy.


Unless stated otherwise, we set $N_{0_{s=1}}=2$, $N_{0_{s=0}}=10$, and $N=20$ BO iterations. 
A shared observation noise variance $\sigma_{\eta}^{2}$ across information sources is estimated via GP marginal log-likelihood \cite{balandat2020botorch}. 
We use $n_{\text{c}}=4$ IS3 corrections and optimize the cost-aware EI in \eqref{eq:caEI} using L-BFGS \cite{Liu1989} to select $(k_{n+1},s_{n+1})$. 
Objective weights and normalization are computed from $10$ Latin Hypercube samples (LHS) \cite{iman1981approach}, yielding $[w_{1},w_{2},w_{3},w_{4}]=[0.02, 0.20, 0.70, 0.20]$, with means and standard deviations used for normalization of objectives. 
We perform a Monte Carlo study with $N_{\text{exper}}=50$ experiments, each initialized with $\mathfrak{D}_{\text{init}}$ drawn from random LHS.



\begin{figure}[!h]
    \centering
    \begin{subfigure}[b]{0.8\columnwidth}
        \centering
        \includegraphics[width=\textwidth, trim=0cm 0cm 0cm 1.6cm, clip]{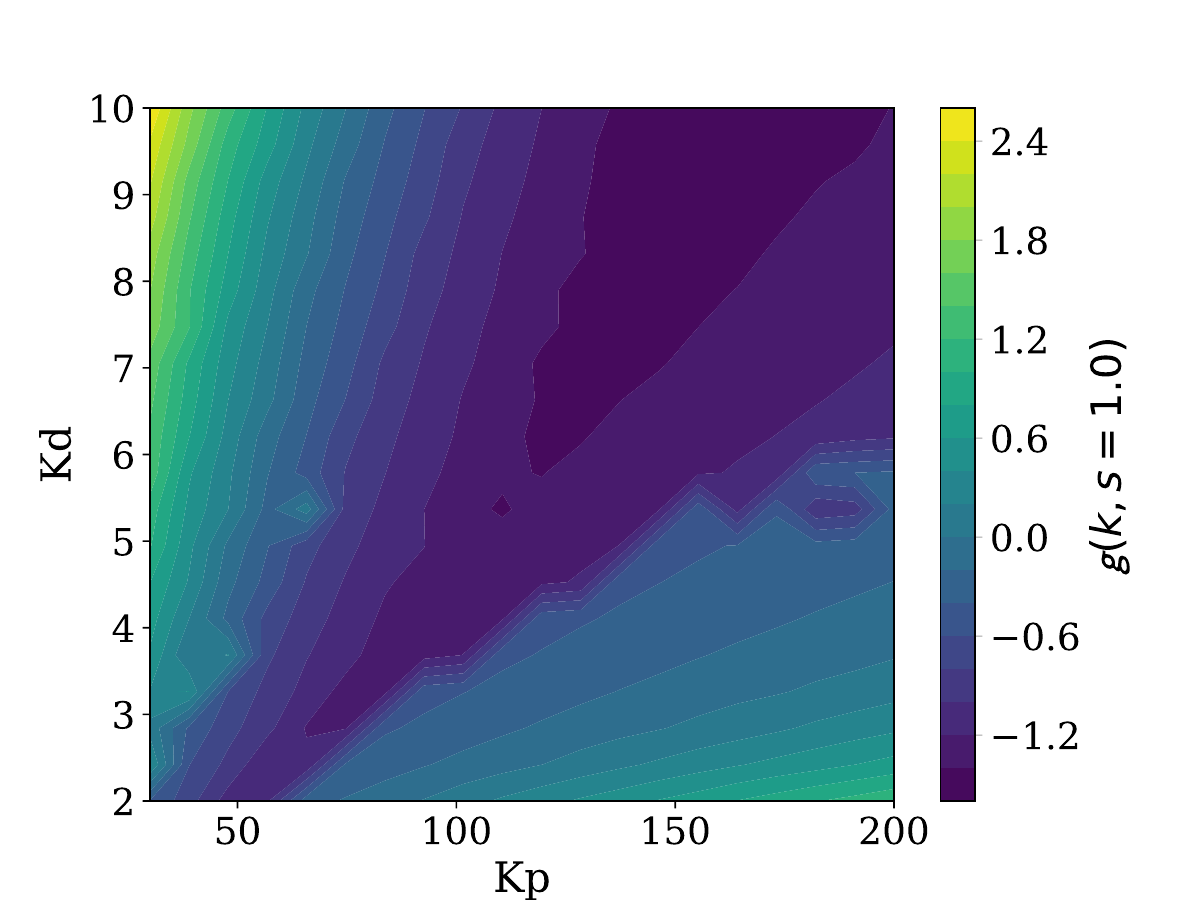}
        
        \label{fig:J_IS1_numeric}
    \end{subfigure}
    \vspace{-0.5\baselineskip}
    \begin{subfigure}[b]{0.8\columnwidth}
        \centering
        \includegraphics[width=\textwidth, trim=0cm 0.4cm 0cm 0.2cm, clip]{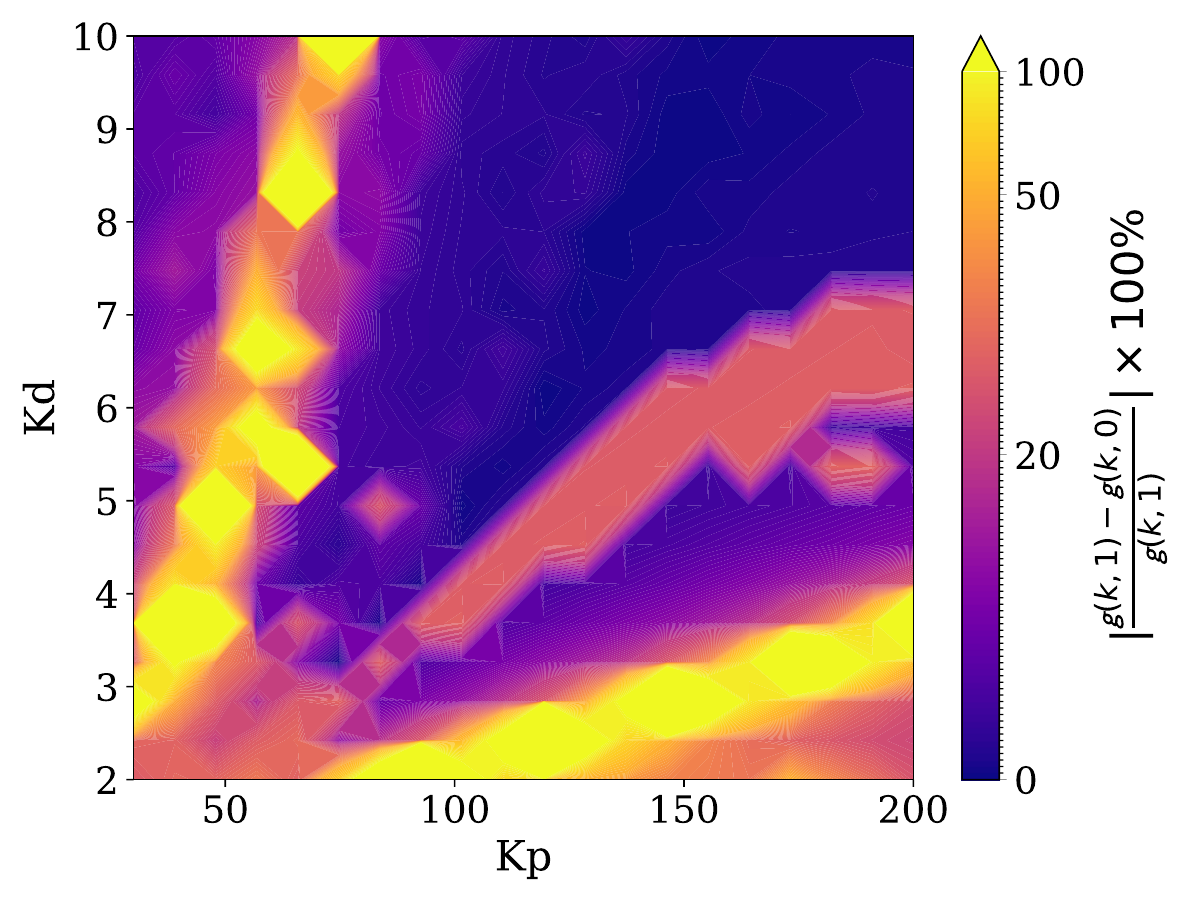}
    
        \label{fig:J_IS2_numeric_ABS_error}
    \end{subfigure}
    \caption{The top panel shows the true objective function given IS1; The bottom panel shows the relative absolute error (in \%) of the estimated objective by IS2}\vspace{-1.\baselineskip}
    \label{fig:IS1_vs_IS2}
\end{figure}

Figure \ref{fig:IS1_vs_IS2} shows the ground-truth objective over $\mathcal{K}$ (top), and the relative prediction error of IS2 (bottom). Errors exceed $100\%$ in some regions, causing IS2 data to mislead the optimizer by distorting the GP surrogate. This motivates correcting IS2 predictions via IS3 and adapting the surrogate and acquisition function to account for variable simulation fidelity.

\vspace{-0.5\baselineskip}
\subsection{Comparison with Baseline Methods}\label{subsec:baseline_comparison}
We compare GMFBO against baseline BO with EI and three MFBO variants: baseline MFBO with trace-aware Knowledge Gradient (taKG) \cite{wu2020practical}, baseline MFBO with our caEI, and a modified MFBO with caEI using the fidelity-aware kernel in \eqref{eq:kernel} without including IS3.
The taKG acquisition weighs expected information gain against sampling cost, which relies strongly on the surrogate GP.
GMFBO explicitly corrects DT bias and guides optimization toward the objective.
In contrast, taKG resembles an active learning strategy that improves surrogate accuracy in informative regions without adapting to biased or shifting fidelities.

We first conduct an ablation study on the size of initial data per information source in $\mathfrak{D}_{\text{init}}$. 
Let $n^{*}$ denote the average IS1 iterations to reach the optimum.
Table \ref{table:initial_dataset_ablation} reports $n^{*}$ with different initial datasets.  
GMFBO performs best with a balanced amount of IS2 data: excess data degrades the surrogate due to IS2 inaccuracies, while insufficient data reduces efficiency.  
The benefit is most pronounced when IS1 samples are scarce.

\setlength{\tabcolsep}{10pt}
\begin{table}[!h]
\centering
\vspace{-0.7\baselineskip}
\setlength\extrarowheight{1.1pt}
\caption{\label{table:initial_dataset_ablation}Required $n^{*}$ given different initial dataset sizes}
\begin{tabular*}{0.99\columnwidth}{@{\extracolsep{\fill}}cc ccc}
\toprule
 &  & \multicolumn{3}{c}{$n^{*}$} \\
\cmidrule(lr){3-5}
${N_{0_{s=1}}}$ & ${N_{0_{s=0}}}$ & EI-BO & MFBO (taKG) & GMFBO \\ \midrule
1  & 10  & 15 & $>21$ & 9 \\
2  & 10  & 22 & $>22$ & 6 \\
3  & 10  & 17 & $>23$ & 10 \\
4  & 10  & 17 & $>24$ & 11 \\
5  & 10  & 15 & $>25$ & 11 \\
10 & 10  & 20 & $>30$ & 15 \\
\cmidrule(lr){1-5}
2  & 5   & 22 & $>22$ & 13 \\
2  & 20  & 20 & $>22$ & 10 \\
2  & 30  & 22 & $>22$ & 10 \\
\bottomrule
\end{tabular*}
\vspace{-0.5\baselineskip}
\end{table}

\begin{figure}[!h] 
\centering
\vspace{-0.6\baselineskip}
\includegraphics[width=.81\columnwidth, trim=1.cm 0.6cm 2.7cm 1.9cm, clip]{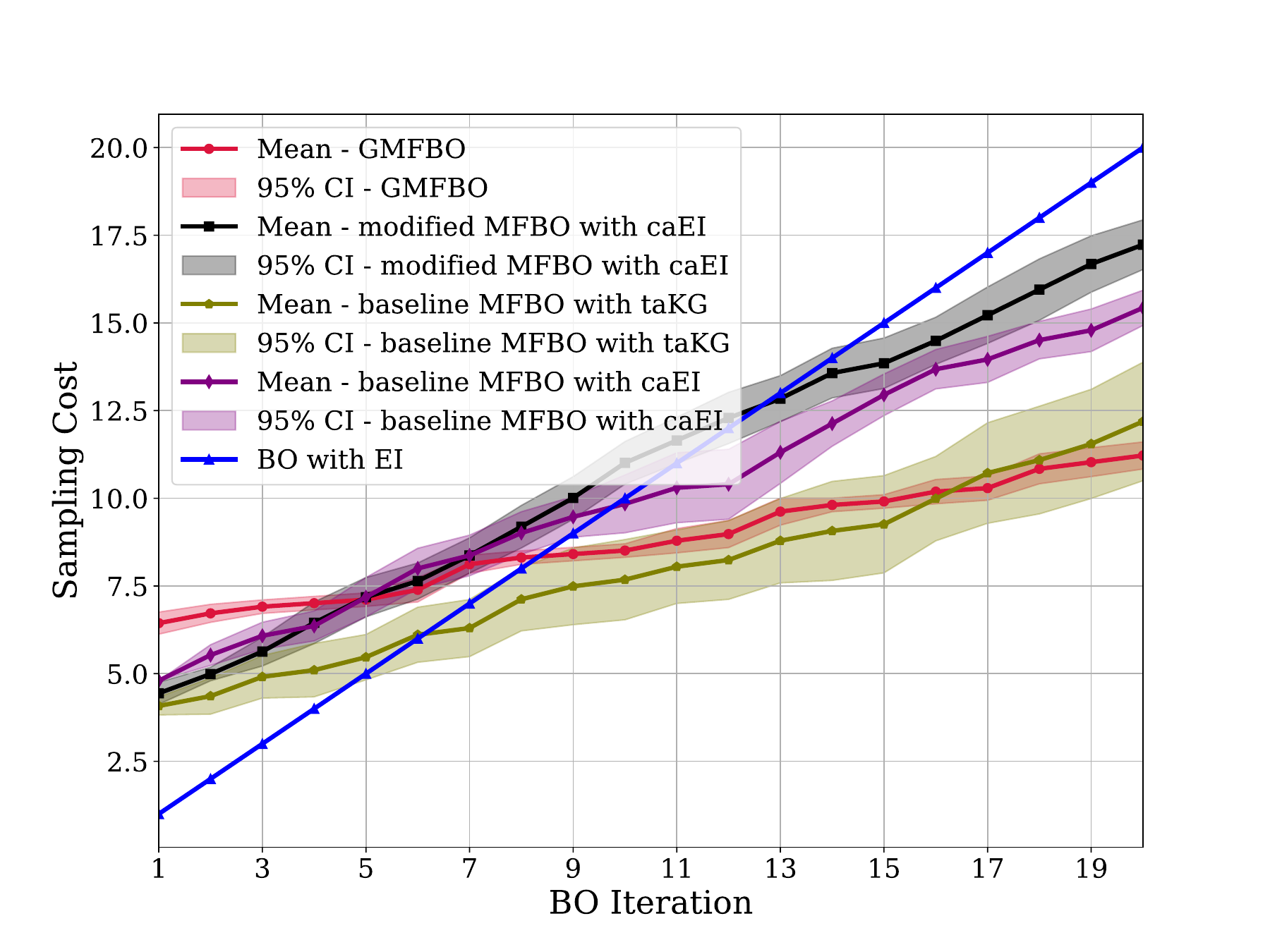}
\caption{Sampling cost over BO iterations, averaged over $50$ Monte Carlo experiments. Shaded regions indicate $95\%$ confidence intervals.
}\vspace{-0.5\baselineskip}
\label{fig:sampling_cost}
\end{figure}

To compare sampling costs across IS selections, we assign a cost of $1$ to each IS1 query and $0.1$ to each IS2 or IS3 query, reflecting the fact that simulations are an order of magnitude cheaper than real-system evaluations while still contributing informative samples.
Figure \ref{fig:sampling_cost} shows the mean and confidence intervals (CI) of these costs for different methods.
GMFBO has a higher initial sampling cost since $N_{0_{s=1}}\cdot n_{\text{c}}=2\cdot 4=8$ IS3 samples are required.
Since IS3 is inexpensive and guides the optimizer to reduce IS1 queries, GMFBO attains a lower cumulative sampling cost than BO and other MFBO methods, whose higher variance results from oversampling inaccurate DT data, inflating the sampling cost.

\begin{figure}[!h] 
\centering
\vspace{-0.5\baselineskip}
\includegraphics[width=.87\columnwidth, trim=1.cm 0.6cm 2.7cm 1.9cm, clip]{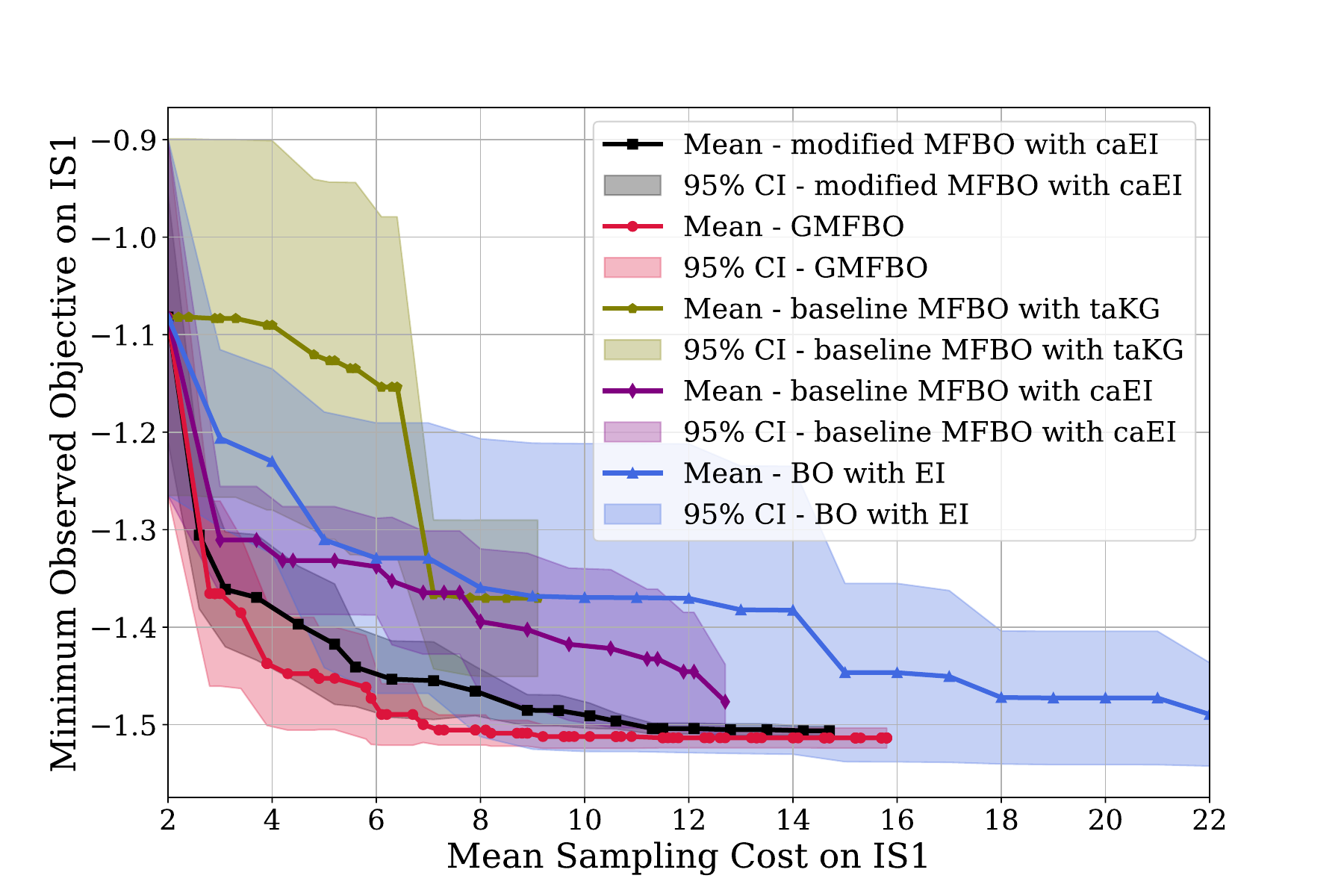}
\caption{Minimum observed IS1 objective vs. mean IS1 sampling cost over $50$ Monte Carlo experiments. Shaded regions denote $95\%$ confidence intervals\label{fig:Jstar_IS1iter}}\vspace{-0.5\baselineskip}
\end{figure}

Figure \ref{fig:Jstar_IS1iter} shows that baseline MFBO with taKG (olive) fails to exploit IS2 and exhibits higher variance than MFBO with caEI (purple) due to poor acquisition. 
Large IS2 errors cause taKG to oversample inaccurate IS2 data, thereby distorting the surrogate model due to the lack of fidelity-aware weighting.
Since taKG estimates the optimum from the posterior GP, its failure arises from IS2’s relative absolute error $\frac{\lvert \mathsf{g}_{\text{c}}(k,1) - \mathsf{g}(k,0) \rvert}{\lvert \mathsf{g}_{\text{c}}(k,1) \rvert}$ visualized in Figure \ref{fig:IS1_vs_IS2}, which exceeds $20\%$ on average.  
Using caEI improves the baseline MFBO to a level comparable to that of BO using EI, although without an apparent reduction in CI.  
The modified MFBO (black) further benefits from the customized kernel, which shrinks variance by adjusting the influence of DT data in the surrogate and preventing optimizer pollution from low-accuracy DR data, leading to improved convergence.  
Our GMFBO, which combines caEI, a customized kernel, and IS3 correction, achieves higher data efficiency by requiring fewer IS1 measurements with reduced and faster-decreasing variability across iterations.  
BO and modified MFBO with caEI need $22$ and $11$ IS1 experiments on average (including initialization) to reach a given threshold, whereas GMFBO requires only $6$, i.e., $72\%$ and $45\%$ fewer, respectively.  
In the best run, GMFBO converges in $2$ experiments, compared to $8$ and $6$ for BO and modified MFBO, yielding $75\%$ and $66\%$ improvements.  
Even in the worst case, GMFBO converges within $8$ experiments, while BO may not converge after $22$, and modified MFBO requires $11$, i.e., improvements of at least $64\%$ and $27\%$.  
These results show that GMFBO outperforms BO and both baseline and modified MFBO methods.

\vspace{-1.2\baselineskip}
\subsection{Adaptation to System Variation}

We study a non-stationary system where the motor friction coefficient in IS1 doubles after the fourth iteration of IS1, while IS2 remains unchanged, thereby reducing its fidelity.

\begin{figure}[!h] 
\centering
\vspace{-0.5\baselineskip}
\includegraphics[width=.87\columnwidth, trim=1.cm 0.5cm 2.7cm 2.2cm, clip]{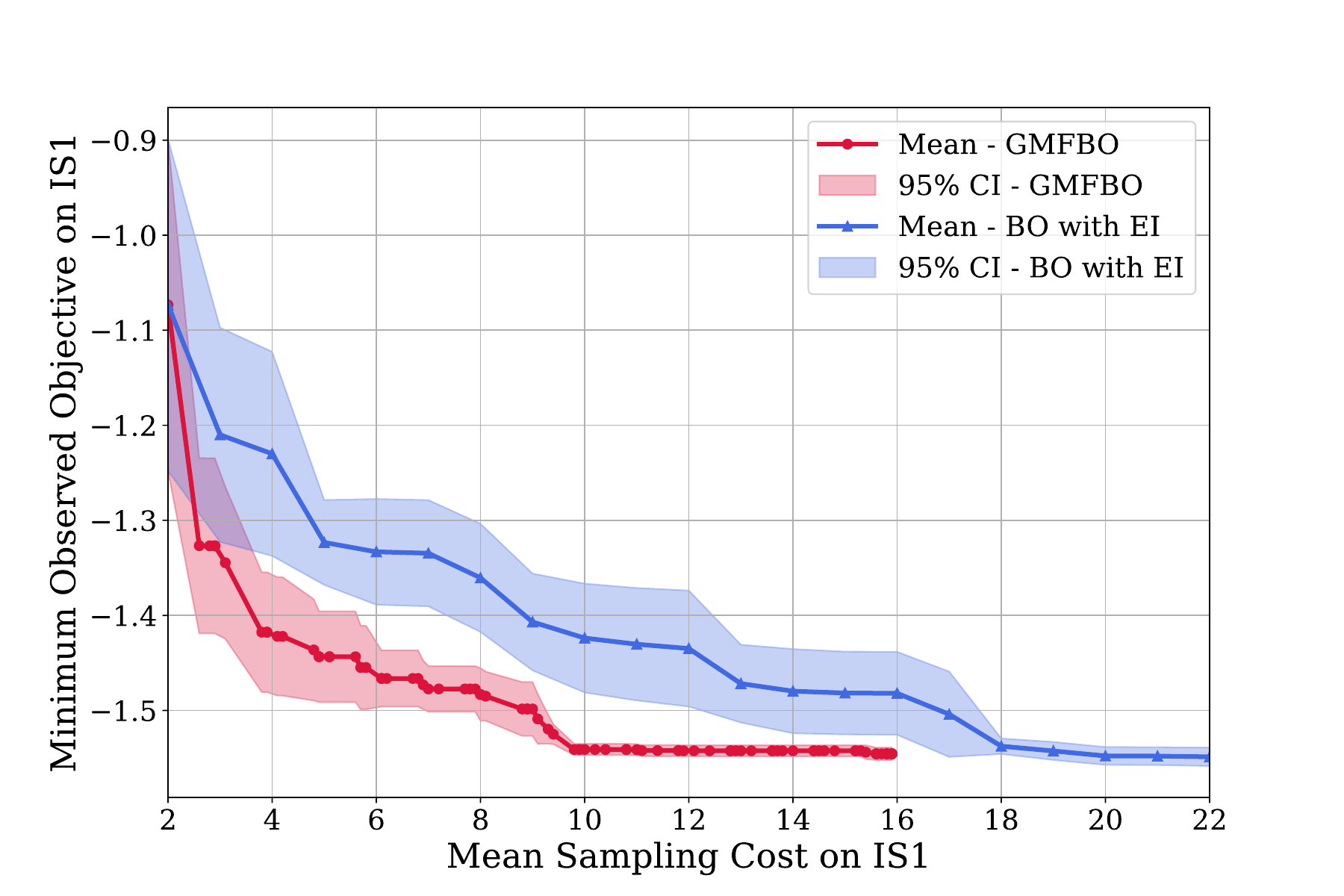}
\caption{Minimum observed IS1 objective vs. mean IS1 cost over $50$ Monte Carlo experiments with varying friction in IS1. Shaded regions denote $95\%$ confidence intervals.\label{fig:numerical_fmin_iter}}\vspace{-0.5\baselineskip}
\end{figure}


As shown in Figure~\ref{fig:numerical_fmin_iter}, GMFBO adapts to the IS1 change and finds the new optimum in $10$ IS1 experiments, compared to over $18$ for BO with EI ($44\%$ improvement).  
The change increases IS2 error, which lowers cross-source correlation and raises IS2 sampling cost, while online IS3 corrections improve the surrogate.  
This allows GMFBO to continue exploiting DT data despite shifting system dynamics.

\vspace{-0.5\baselineskip}
\section{Hardware Validation}\label{sec:Exper_Results}
\begin{figure}[!h] 
\centering
\vspace{-.5\baselineskip}
\includegraphics[width=.85\columnwidth, trim=3.9cm 0.1cm 3.9cm 0cm, clip]{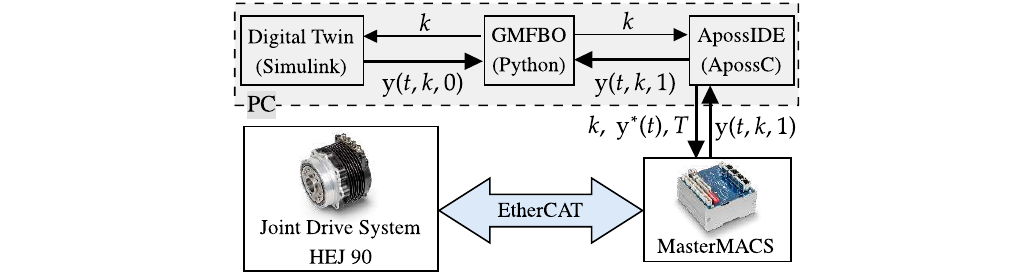}
\caption{Real system hardware setup used to evaluate GMFBO\label{fig:real_system}}
\vspace{-0.5\baselineskip}
\end{figure}

We validate data efficiency and generalization on the Maxon HEJ 90 drive system (Figure~\ref{fig:real_system}), with IS1 as the real hardware and IS2 as the DT (corresponding to IS1 in Section \ref{sec:Numerical_Analysis}).
The integrated EPOS4 controller employs a hierarchical structure comprising current, position, velocity, and torque loops, all of which are accessed via EtherCAT.
A custom MasterMACS–ApossIDE integration enables real-time acquisition of encoder positions and control inputs at $1$\,kHz \cite{maxon2025robotics}.

Since the objective metrics (overshoot, rise, settling, and transient times) differ in scale and estimation accuracy, we normalize them and assign weights that reflect both application relevance and IS2 reliability.  
Weights can be adapted to task priorities or evolving DT accuracy; in our case,  
$w_{1} = 0.01, \; w_{2} = 0.09, \; w_{3} = 0.82, \; w_{4} = 0.09$.  
\begin{figure}[!h] 
\centering
\vspace{-0.5\baselineskip}
\includegraphics[width=.9\columnwidth, trim=1.1cm 0.3cm 1.6cm 1.9cm, clip]{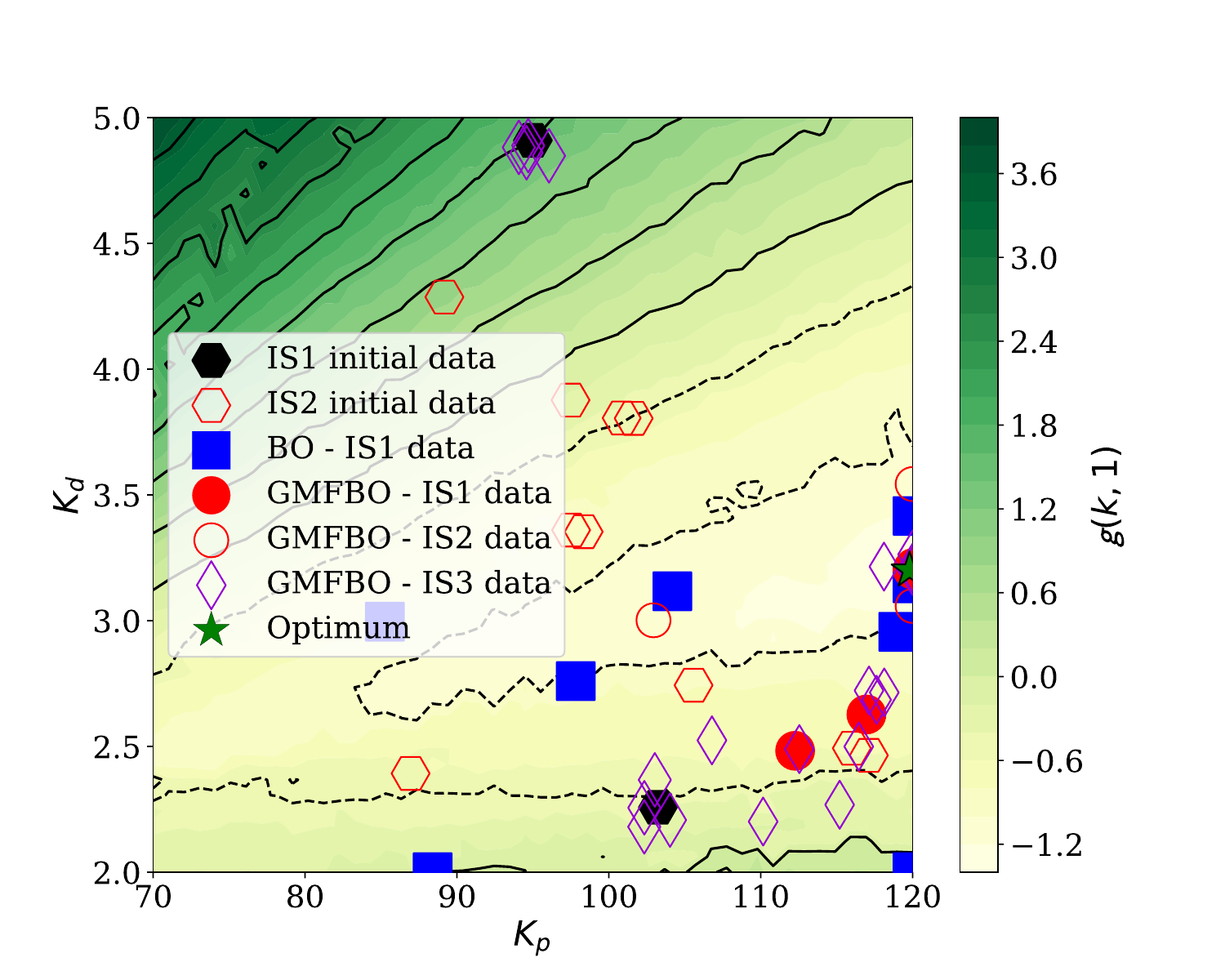}
\caption{Sampled data points of GMFBO and baseline BO with EI in a single experiment.  
Iterations are shown until the minimum objective approaches the global optimum (star).  
Contours depict the true objective for the real hardware.}\vspace{-0.5\baselineskip}
\label{fig:real_iteration_data_points}
\end{figure}

Figure \ref{fig:real_iteration_data_points} shows sampled data points per method in a single experiment until the optimum is found.  
Baseline BO relies heavily on costly IS1 measurements, whereas GMFBO shifts part of the exploration and exploitation to cheaper IS2 and IS3, reducing IS1 usage.

\begin{figure}[!h] 
\centering
\vspace{0.3\baselineskip}
\includegraphics[width=.85\columnwidth, trim=0cm 0cm 0cm 0.1cm, clip]{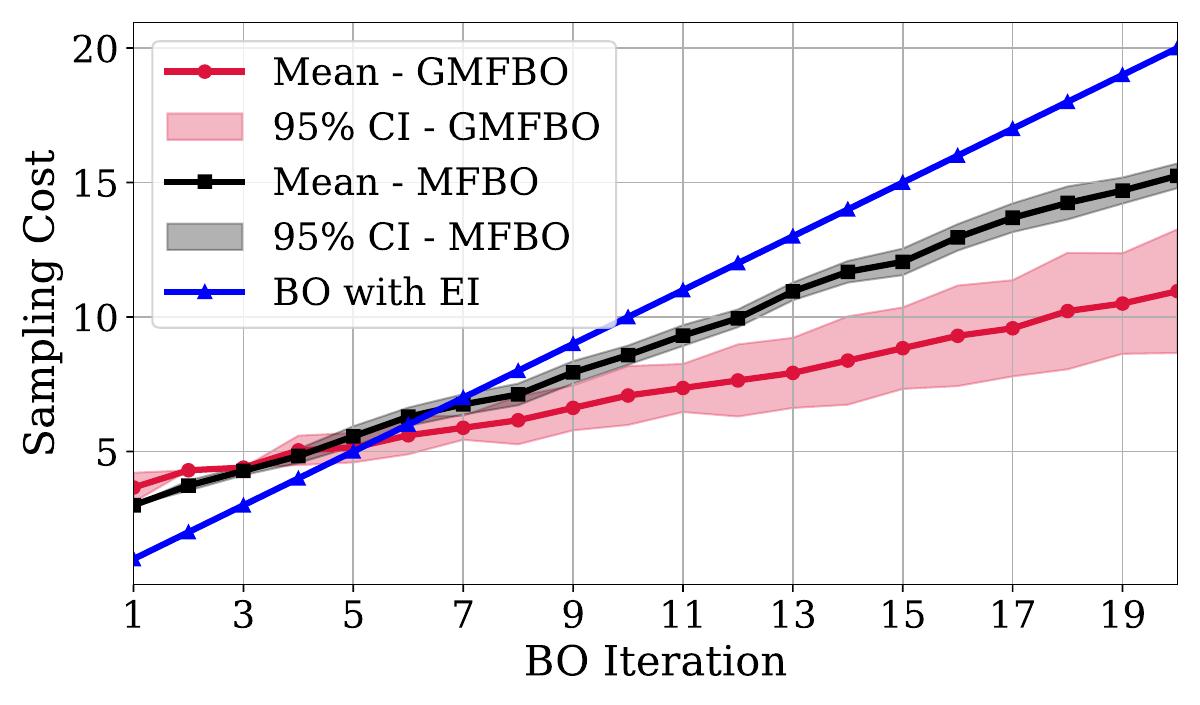}
\caption{Sampling cost over BO iterations from $20$ Monte Carlo experiments. Shaded regions denote $95\%$ confidence intervals.}
\vspace{-1.\baselineskip}
\label{fig:sampling_cost_real_system}
\end{figure}

We conduct a Monte Carlo analysis with $N_{\text{exper}}=20$ experiments.
Figure \ref{fig:sampling_cost_real_system} shows that our GMFBO method incurs lower total sampling cost than both BO and MFBO over $N=20$ BO iterations.  
We also observe variability in the sampling cost for GMFBO and MFBO, stemming from the dynamic selection between two information sources guided by our acquisition function.  
Additionally, the confidence interval for GMFBO’s sampling cost is wider than that of MFBO, reflecting greater cost variability due to IS3's reliance on real system measurements.

\begin{figure}[!h] 
\centering
\vspace{-0.3\baselineskip}
\includegraphics[width=.99\columnwidth, trim=.81cm 0cm 0.8cm 1.2cm, clip]{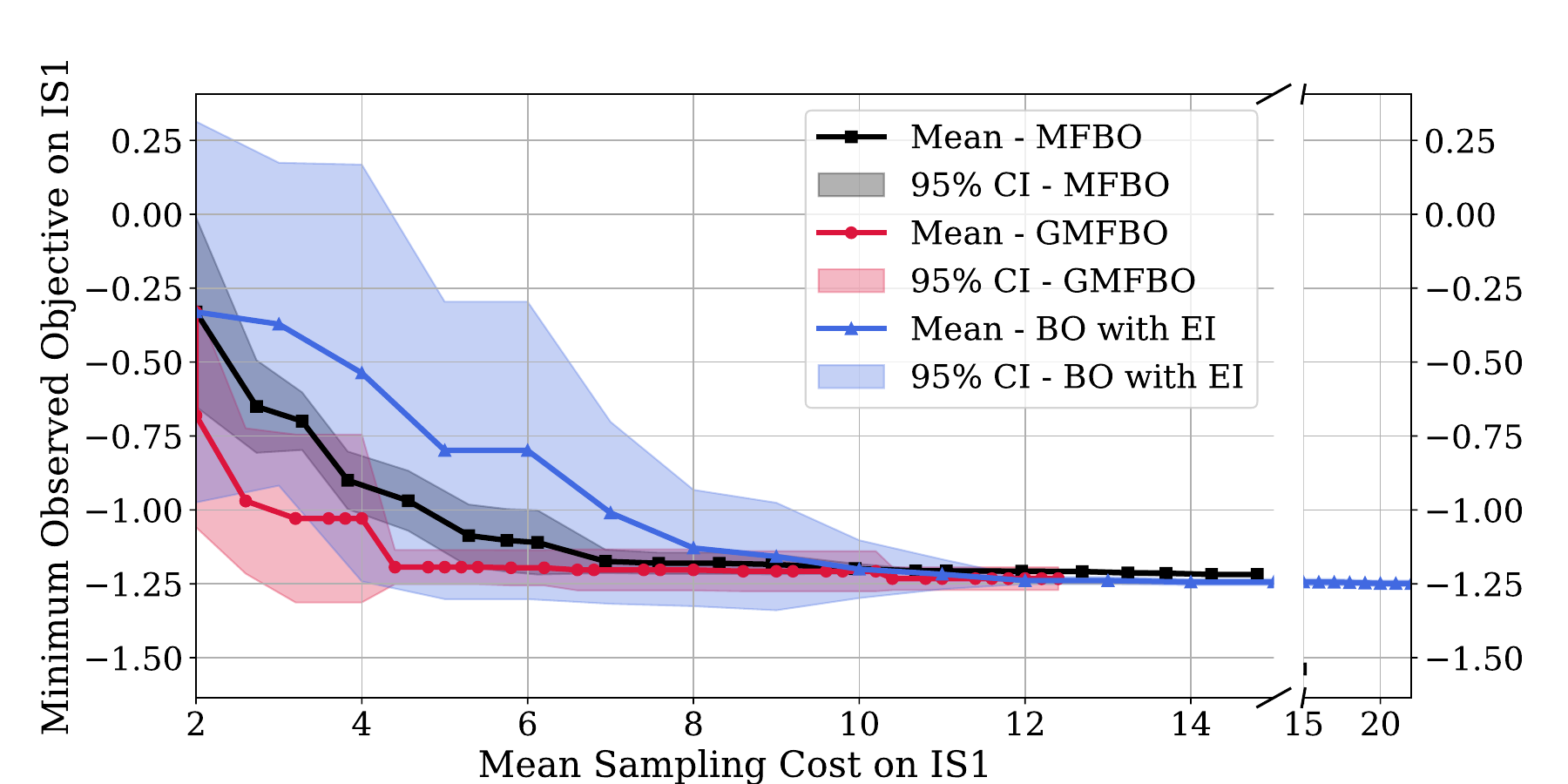}
\caption{Minimum observed IS1 objective vs. mean IS1 cost on the real system from $20$ Monte Carlo experiments.
Shaded denotes $95\%$ confidence interval.\label{fig:Jstar_IS1iter_real}}
\vspace{-.5\baselineskip}
\end{figure}

Figure \ref{fig:Jstar_IS1iter_real} shows that GMFBO requires fewer real-system experiments on average, with a smaller and faster-shrinking confidence interval than other methods.  
BO and MFBO need $9$ and $7$ experiments, respectively, while GMFBO achieves the same performance in only $5$, yielding $44\%$ and $28\%$ improvements.  
Among the experiments, the best GMFBO converges in $2$ experiments (vs. $6$ for BO and $4$ for MFBO, $67\%$ and $50\%$ improvements), and the worst in $6$ (vs. $13$ and $8$, $54\%$ and $25\%$ improvements).  

\vspace{-0.5\baselineskip}
\section{Conclusion}\label{sec:Conclusion}
This paper introduces a guided multi-fidelity Bayesian optimization (GMFBO) framework for data-efficient controller tuning, which systematically incorporates simulation fidelity into the surrogate model.  
Our main contributions are: (i) a customized multi-source kernel that adaptively scales cross-fidelity correlations based on estimated digital-twin accuracy, preventing low-fidelity data from degrading the surrogate; (ii) an adaptive cost-aware acquisition function that balances sampling cost, fidelity, and expected improvement for efficient query selection; and (iii) integration of real-data corrections to maintain robustness under model mismatch and adapt to varying simulation fidelities.  

Extensive numerical studies have shown that GMFBO consistently outperforms existing multi-fidelity methods in terms of convergence speed and sample efficiency.  
On robotic drive hardware, GMFBO achieved over $28\%$ improvement in optimization efficiency compared to baselines.  

These results demonstrate the potential of GMFBO to accelerate learning in systems with costly experiments and imperfect models.
Future work will extend the framework to settings where fidelity varies with the controller gains (input-dependent) or the system’s operating conditions (state-dependent, e.g., velocity or load).
Other promising directions are closed-loop, data-efficient digital twin identification and robustness under non-convex and discontinuous objectives.

\vspace{-0.5\baselineskip}



\bibliographystyle{IEEEtran}
\bibliography{main}


\end{document}